\documentclass[letterpaper]{article} % DO NOT CHANGE THIS
\usepackage{aaai2027}  % DO NOT CHANGE THIS
\usepackage[hyphens]{url}  % DO NOT CHANGE THIS
\usepackage{graphicx} % DO NOT CHANGE THIS
\usepackage{natbib}  % DO NOT CHANGE THIS AND DO NOT ADD ANY OPTIONS TO IT
\usepackage{caption} % DO NOT CHANGE THIS AND DO NOT ADD ANY OPTIONS TO IT
\usepackage{algorithm}
\usepackage{algorithmic}

\usepackage{newfloat}
\usepackage{listings}
\DeclareCaptionStyle{ruled}{labelfont=normalfont,labelsep=colon,strut=off} % DO NOT CHANGE THIS
\floatstyle{ruled}
\newfloat{listing}{tb}{lst}{}
\floatname{listing}{Listing}

\usepackage{booktabs}

\usepackage{enumitem}
\usepackage{amsmath}
\usepackage{amssymb}
\usepackage{dsfont}
\usepackage{multirow}
\usepackage{threeparttable}

\usepackage{makecell}

\title{DySink: Dynamic Frame Sinks for Autoregressive Long Video Generation}
\author{
    Bo Ye\textmd{\textsuperscript{1,2,3}}\textmd{,}
  ~Xinyu Cui\textmd{\textsuperscript{4}}\textmd{,}
  ~Jian Zhao\textmd{\textsuperscript{3,5}}\textmd{,}
  ~Tong Wei\textmd{\textsuperscript{1,2}}\textmd{,} 
  ~Min-Ling Zhang\textmd{\textsuperscript{1,2}}
}
\affiliations{
\textsuperscript{1}School of Computer Science and Engineering, Southeast University\\ \textsuperscript{2}Key Lab. of Computer Network and Information Integration, Southeast University\\ \textsuperscript{3}Zhongguancun Academy\\
\textsuperscript{4}Institute of Automation, CAS\\
\textsuperscript{5}Zhongguancun Institute of Artificial Intelligence
}

\begin{document}

\maketitle

\begin{abstract}
Autoregressive long video generation often adopts bounded-memory streaming for efficiency, typically combining local windows for short-term continuity with static early-frame sinks as long-range anchors. However, this fixed allocation keeps early frames cached even when the current visual state has substantially diverged from them, while discarding potentially more relevant intermediate history. As a result, the retained long-range context may become less adaptive and bias generation toward outdated cues; in severe cases, RoPE-induced phase re-alignment can homogenize inter-head attention and cause sink collapse, where content regresses toward sink frames. We propose \emph{DySink}, a retrieval-based framework that maintains a compact memory bank and selects visually relevant historical frames as dynamic frame sinks. DySink couples adaptive retrieval with a sink anomaly gate that filters retrieved context exhibiting excessive inter-head consensus, an attention pattern associated with sink collapse. Experiments on 50--100-second videos show that DySink achieves the highest measured temporal quality among the evaluated autoregressive baselines, while retaining competitive text alignment and framewise quality. The code is available at ~\url{https://github.com/yebo0216best/DySink}.
\end{abstract}

\section{Introduction}

Diffusion-based video generation has advanced rapidly, with systems such as Sora~\citep{openai2024sora}, Wan~\citep{wan2025wan}, and Seedance~\citep{seedance2026seedance} demonstrating impressive visual fidelity and motion realism. Despite this progress, extending generation from short videos (e.g., 5--10 seconds) to long-horizon videos remains challenging. A major bottleneck lies in the quadratic cost of bidirectional diffusion transformers~\citep{dit} over spatiotemporal tokens, which makes direct long-horizon video generation computationally prohibitive. To improve scalability, recent studies have turned to autoregressive streaming paradigms, where videos are generated sequentially with causal attention and KV caching. This formulation enables scalable long-horizon generation with bounded memory. Representative methods, including CausVid~\citep{yin2025causvid}, Self-Forcing~\citep{huang2025self}, Self-Forcing++~\citep{cui2025self}, Rolling-Forcing~\citep{liu2025rolling}, and LongLive~\citep{yang2025longlive}, progressively extend generation horizons to minutes by combining streaming training with fixed-size context. Under this setting, a central design question is how to allocate the limited historical cache so that the model can maintain temporal continuity and long-range consistency. Recent methods~\citep{liu2025rolling,yang2025longlive} commonly address this issue by combining sliding-window attention with \emph{frame-level attention sinks} (hereafter, \emph{frame sinks})~\citep{xiao2023efficient}. In this design, a local window retains recent frames for short-term continuity, while several early frames are cached as global anchors for long-term
consistency.

\begin{figure*}[t]
% \vspace{-1pt}
\begin{center}
\includegraphics[width=0.98\linewidth]{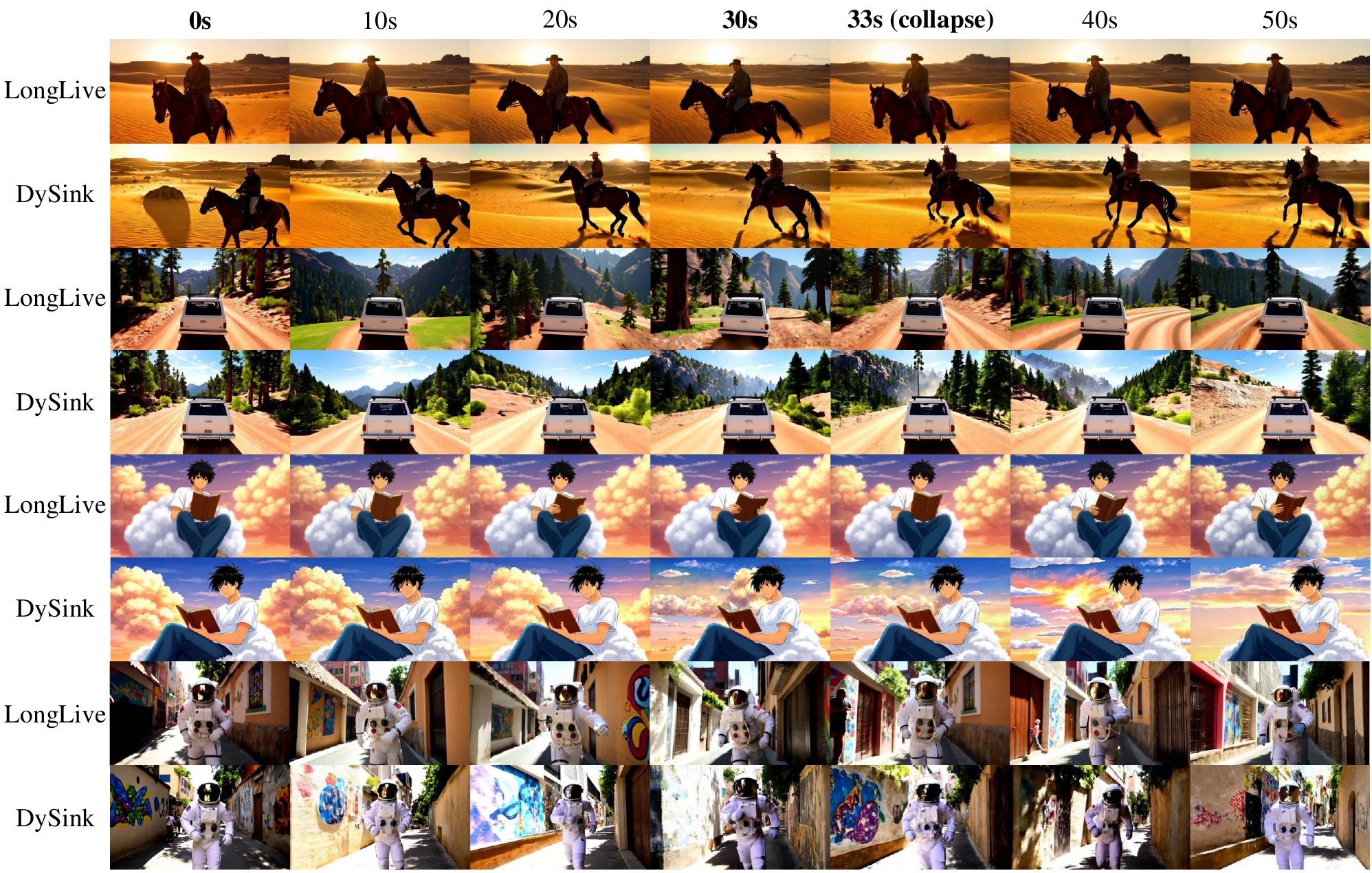}
\end{center}
\caption{
\textbf{Motivating comparison.}
We compare the static-sink baseline LongLive~\citep{yang2025longlive} with DySink. LongLive tends to revisit early visual states, whereas DySink supports continued visual evolution over long rollouts.
}
\label{fig:motivation}
\end{figure*}

Despite their empirical success, static frame sinks impose a fixed memory allocation that may become suboptimal in long rollouts. Under a fixed-size memory, sliding-window attention may discard intermediate frames that better match later visual states, while early sink frames remain persistently cached even after the visual state has substantially evolved. As a result, the model may rely on outdated anchors rather than more relevant historical frames, biasing generation toward misaligned visual cues~\citep{yang2026stableworld}. This suggests that the issue lies not in long-range conditioning itself, but in the static selection of long-range context: early-frame anchors can provide useful stabilizing cues, yet their fixed allocation becomes less adaptive as generation evolves. In the qualitative examples shown in Figure~\ref{fig:motivation}, LongLive revisits early visual cues, whereas DySink exhibits more substantial visual evolution while preserving coherence. In more severe cases, \citet{cui2026lol} identify sink collapse, where generated content repeatedly regresses toward sink frames, producing abrupt scene resets and cyclic motion patterns. Their analysis attributes this collapse to phase re-alignment under RoPE~\citep{su2024roformer} and inter-head attention homogenization, where many attention heads simultaneously assign high weights to sink frames.

In this work, we propose DySink for adaptive long-range memory selection in autoregressive long video generation. Rather than keeping the earliest frames as persistent anchors, DySink maintains a compact memory bank and retrieves historical frames that are visually relevant to the current generation context. These retrieved frames serve as dynamic frame sinks, providing long-range conditioning without permanently relying on fixed early-frame anchors. To reduce sink-collapse-prone attention patterns, DySink further couples retrieval with a lightweight per-layer sink anomaly gate. The gate is motivated by the observation that sink collapse is associated not with a single attention head, but with excessive inter-head consensus, where many heads simultaneously over-attend to retrieved long-range context. Instead of modifying RoPE with multi-head RoPE jitter~\citep{cui2026lol}, DySink preserves the original positional encoding structure and uses abnormal consensus over retrieved context as an indicator that the selected memory may be collapse-prone. When such consensus is observed for a retrieved block, the gate suppresses that block from the layer-wise KV context while retaining other retrieved blocks; the layer falls back to the sliding-window context only when all retrieved blocks are filtered out. This design decouples which history to reuse from when to trust it, enabling adaptive long-range conditioning while controlling collapse-prone retrieved context. Our contributions are summarized as follows:
\begin{itemize}[leftmargin=*]

\item We introduce \textbf{DySink}, a streaming framework that replaces
fixed early-frame anchors with a novelty-aware memory bank and retrieves
relevant historical KV blocks under a fixed active-context budget.

\item DySink decouples \emph{what history to retrieve} from \emph{when to trust
it}. A lightweight block-wise anomaly gate filters retrieved context exhibiting
excessive inter-head consensus without modifying RoPE or attention logits.

\item Experiments on 50--100s videos show that DySink achieves the highest
temporal quality among the autoregressive
baselines while retaining competitive text alignment and framewise quality.

\end{itemize}

\section{Related Work}
\label{sec:related}

\subsection{Autoregressive Video Diffusion}

Autoregressive video diffusion has emerged as a practical paradigm for streaming video generation, where future frames or chunks are generated causally from previously synthesized content. CausVid~\citep{yin2025causvid} distills bidirectional video diffusion models into few-step causal generators with KV caching, while Self-Forcing~\citep{huang2025self} reduces the train--test gap by rolling out the model on its own predictions during training. Subsequent works further improve this paradigm through stronger rollout training, distillation, or optimization objectives: Self-Forcing++~\citep{cui2025self} and Rolling-Forcing~\citep{liu2025rolling} improve long-horizon stability, Reward Forcing~\citep{lu2025reward} enhances motion dynamics with reward-weighted distribution matching, and Resampling Forcing~\citep{guo2025end} studies teacher-free training with self-resampled histories. Helios~\citep{yuan2026helios} further explores real-time long-video generation at larger model scale. These methods mainly address how to obtain robust and efficient autoregressive generators. Causal Forcing~\citep{zhu2026causal} further improves autoregressive distillation by using an autoregressive teacher for causal ODE initialization, addressing the architectural gap between bidirectional teachers and causal students. In this work, our focus is complementary: given such a generator, DySink studies how its historical context should be selected and reused during long rollouts.

\subsection{Long Video Generation}

Long video generation requires maintaining scene identity and temporal coherence across extended rollouts while allowing the visual content to evolve over time. A common solution is to restrict attention to a sliding window, which enables scalable generation but discards distant context. To compensate, frame sinks have been adopted as persistent global anchors. LongLive~\citep{yang2025longlive} and Rolling Forcing~\citep{liu2025rolling} retain early frames as static frame sinks. LoL~\citep{cui2026lol} further identifies a sink-collapse failure mode in ultra-long autoregressive generation, where generated content repeatedly regresses toward sink frames. Related methods improve long-range memory by enlarging, updating, compressing, or structuring historical context: Deep Forcing~\citep{yi2025deep} uses deep sinks with training-free KV compression, Reward Forcing~\citep{lu2025reward} updates sink states with EMA, Relax Forcing~\citep{zhao2026relax} decomposes history into Sink, Tail, and History regions, VideoSSM~\citep{yu2025videossm} maintains a SSM-based compressed global memory, and Pretraining Frame Preservation~\citep{zhang2025pretraining} learns lightweight history embeddings. For interactive or narrative generation, Anchor Forcing~\citep{yang2026anchor} designs anchor-guided re-caching for prompt switches. Context-as-Memory~\citep{yu2025context} treats historical frames as memory and retrieves relevant context according to camera-trajectory-based FOV overlap, enabling scene-consistent interactive long-video generation. MemFlow~\citep{ji2025memflow} retrieves prompt-relevant historical cues. Unlike these methods that rely on static, compressed, rule-updated, camera-guided, or prompt-relevant memories, DySink retrieves visually relevant historical KV states from a dynamic memory bank and further controls collapse-prone retrieved context with the sink anomaly gate.

\begin{figure*}[t]
\begin{center}
\includegraphics[width=0.98\linewidth]{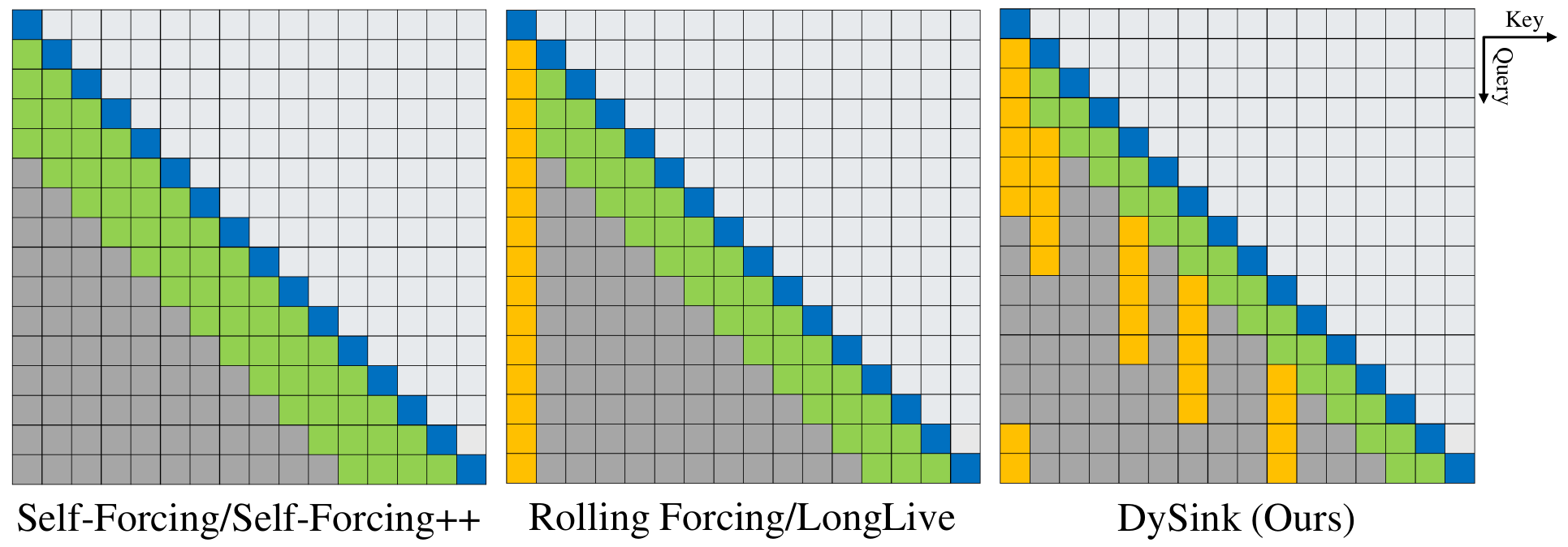}
\end{center}
\caption{
\textbf{Comparison of attention patterns for autoregressive long video generation.}
Blue, green, yellow, and gray cells denote current frames, local-window frames,
long-range anchor frames, and inactive historical frames, respectively.
Self-Forcing~\citep{huang2025self} and Self-Forcing++~\citep{cui2025self}
use only local-window frames, causing distant history to be discarded.
Rolling Forcing~\citep{liu2025rolling} and LongLive~\citep{yang2025longlive}
introduce static frame sinks for long-range consistency. DySink retrieves relevant historical
frames as frame sinks.
}
\label{fig:dysink}
\end{figure*}

\section{Method}
\label{sec:method}

\subsection{Motivation}

Autoregressive long-video generation requires semantic and visual consistency
over extended frame sequences, beyond the effective receptive field of
sliding-window attention. Recent methods~\citep{yang2025longlive,
liu2025rolling} address this limitation using frame sinks, where early frames
are permanently cached as global anchors. However, as generation progresses,
the current visual state may substantially diverge from these early-frame
anchors. Retaining such outdated anchors can create a conditioning mismatch
and bias the model toward reproducing sink-frame
attributes~\citep{yang2026stableworld}. Consequently, excessive reliance on
static sinks may reduce adaptability to scene evolution and motion changes and,
in severe cases, lead to sink-collapse-like failures such as abrupt scene
resets or cyclic motion patterns. 

A natural alternative is to select historical context adaptively rather than
treating the earliest frames as permanent anchors. When the retrieved history
is visually aligned with the current generation window, it can provide
non-local structural cues while reducing the risk of imposing outdated visual
priors. Motivated by this observation, we propose DySink, a retrieval-based
alternative to static frame sinks. DySink maintains a compact memory bank of
historical blocks, indexed by visual descriptors, and retrieves relevant KV
states according to the current local context. This content-adaptive mechanism
preserves long-range conditioning without permanently relying on early frames,
with the goal of reducing conditioning mismatch. Adaptive retrieval reduces dependence on fixed early-frame anchors, but it
does not guarantee that every selected historical block remains reliable. Any long-range block, including a
dynamically retrieved one, may still induce collapse-prone attention patterns
because of long-range positional effects. Prior analysis associates sink
collapse with RoPE phase re-alignment and excessive inter-head consensus~\citep{cui2026lol}.

While LoL~\citep{cui2026lol} mitigates this phenomenon by applying multi-head
RoPE jitter, DySink adopts a memory-control strategy. Our sink anomaly gate
uses excessive inter-head consensus as a practical warning signal and
suppresses only the affected retrieved block at the corresponding layer. Other
retrieved blocks remain available, and the layer falls back to the local window
only when all retrieved blocks are filtered out. Thus, retrieval determines
\emph{what history to reuse}, whereas the gate determines \emph{when that
history should be trusted}, without modifying RoPE or directly perturbing the
attention logits.

\subsection{The DySink Framework}

% DySink implements this two-part memory-control strategy using a retrieval-based
% memory bank and a per-layer anomaly gate. For efficient storage and retrieval,
% historical frames are organized into temporal blocks. Existing autoregressive methods allocate the bounded historical context differently. Local-window methods retain only recent frames, while static-sink methods additionally preserve several early frames as long-range anchors. As illustrated in Figure~\ref{fig:dysink}, DySink instead retrieves visually relevant non-local historical blocks while retaining a recent local window. At each autoregressive step $j$, the model generates one video block while conditioning every DiT layer on a fixed-budget historical KV context.

At each autoregressive step $j$, the model generates one video block while
conditioning each DiT layer on a fixed-budget historical KV context organized
into temporal blocks. Local-window methods allocate this budget entirely to
recent blocks, whereas static-sink methods reserve part of it for fixed early
blocks. As illustrated in Figure~\ref{fig:dysink}, DySink instead combines a
recent local window with visually relevant non-local blocks retrieved from a
memory bank. A per-layer anomaly gate further filters collapse-prone retrieved
blocks before attention.

\paragraph{Memory bank construction.}

DySink maintains a memory bank $\mathcal{M}$ over previously generated video blocks, where each entry is defined as $(\mathbf{f}_i, \mathcal{K}_i)$.
Here $\mathbf{f}_i \in \mathbb{R}^d$ is a compact visual descriptor of block $i$, and
$\mathcal{K}_i =
\{(\mathbf{K}_i^{(\ell)}, \mathbf{V}_i^{(\ell)})\}_{\ell=1}^{N_{\mathrm{layer}}}$
denotes the layer-wise key--value caches. To construct $\mathbf{f}_i$, the latent frames of block $i$ are first decoded into pixel space and then encoded by a frozen visual encoder. The resulting frame-wise features are subsequently aggregated via mean pooling and $\ell_2$ normalization~\citep{bolya2025pe}:
\begin{equation}
    \mathbf{f}_i
    = \mathrm{Norm}\!\left(
        \frac{1}{N_i} \sum_{r=1}^{N_i}
        \mathrm{VisualEncoder}(v_i^{(r)})
    \right),
\label{eq:ve}
\end{equation}
where $N_i$ is the total number of pixel frames $v_i^{(r)}$ in block $i$. This block-level representation aggregates short-term visual content while yielding discriminative features for retrieval. We adopt a visual-only indexing scheme, as visual similarity provides a simple and effective signal for temporal coherence.

To maintain a compact and high-quality memory bank, we use a novelty-aware update strategy. The first block is inserted to initialize the memory bank, but is not privileged during retrieval. Thereafter, a candidate block is admitted only if it is sufficiently dissimilar to existing blocks: $\max_{m \in \mathcal{M}} \cos(\mathbf{f}_i, \mathbf{f}_m) \leq \tau_{\mathrm{dedup}}$. This criterion suppresses redundant storage and mitigates the practical growth
of the memory bank. When near-duplicate blocks arise, earlier blocks are
preferentially retained, as they are generally less affected by long-term error
accumulation. 

Memory blocks not used at the current step are kept in CPU memory. At each
generation step, only up to $k$ retrieved blocks and a fixed-budget local window
are transferred to the GPU. Therefore, although novelty-aware deduplication
reduces the storage growth of the historical memory bank, it is the active
context and the associated per-step attention footprint that remain fixed in
size, independent of the generated video length.

% This criterion suppresses redundant storage and prevents the memory bank from being dominated by similar content. 

% Moreover, when near-duplicate blocks arise, earlier blocks are preferentially maintained, as they are generally less corrupted by long-term error accumulation. Finally, memory blocks not used by the current step are offloaded to CPU memory, so the GPU-resident historical context remains bounded by the local window and the retrieved blocks.

\paragraph{Memory retrieval and injection.}
Let \(W_j\) denote the set of block indices within the local sliding window immediately preceding block \(j\). For each block \(w \in W_j\), we compute its descriptor \(f_w\) using the same procedure as in Eq.~\ref{eq:ve}. Given a target block $j$, DySink retrieves relevant historical context from the memory bank $\mathcal{M}$ in a content-adaptive manner. During retrieval, we exclude memory blocks that already fall inside the current sliding window to avoid duplicating local context after KV concatenation. We aggregate blocks within sliding window $\mathcal{W}_j$ as retrieval queries and compute the relevance score of a memory entry $i$ as
\begin{equation}
    s(i,j) = \frac{1}{|\mathcal{W}_j|} \sum_{w \in \mathcal{W}_j} \cos(\mathbf{f}_w, \mathbf{f}_i),
    \label{eq:multi-query}
\end{equation}
where $\cos(\cdot,\cdot)$ denotes cosine similarity. This formulation conditions retrieval on the recent context while smoothing transient frame-level variations. We then select the top-$k$ eligible memory blocks with the highest relevance scores. The corresponding KV caches of the selected blocks are transferred to GPU memory. For each DiT layer $\ell$, the retrieved KV blocks and the local-window KV cache together form the candidate historical context before gating. The anomaly gate described next operates only on the retrieved blocks, while the
local-window context is always retained.

% For each DiT layer $\ell$, the retrieved and local KV caches are concatenated to form the candidate
% attention context before gating:

% % \begin{equation}
% % \mathbf{K}^{(j,\ell)}_{\mathrm{attn}}
% % =
% % \mathrm{Concat}_{\mathrm{token}}
% % \!\left(
% % \mathbf{K}^{(j,\ell)}_{\mathrm{ret}},
% % \mathbf{K}^{(j,\ell)}_{\mathrm{loc}}
% % \right),
% % \qquad
% % \mathbf{V}^{(j,\ell)}_{\mathrm{attn}}
% % =
% % \mathrm{Concat}_{\mathrm{token}}
% % \!\left(
% % \mathbf{V}^{(j,\ell)}_{\mathrm{ret}},
% % \mathbf{V}^{(j,\ell)}_{\mathrm{loc}}
% % \right).
% % \label{eq:kv-compose}
% % \end{equation}

% \begin{equation}
% K_{\mathrm{cand}}^{(j,\ell)}
% =
% \operatorname{Concat}_{\mathrm{token}}
% \left(
% K_{\mathrm{ret}}^{(j,\ell)},
% K_{\mathrm{loc}}^{(j,\ell)}
% \right),
% \label{eq:kv-compose}
% \end{equation}
Retrieved KV states retain their original RoPE positions without re-indexing. When fewer
than $k$ eligible memory blocks are available, we use all available blocks and
expand the local window by the missing number of blocks to preserve the context
budget. Unlike prior approaches that rely on static sinks, DySink dynamically retrieves visually relevant historical context from the memory bank, thereby reducing conditioning mismatch.

\begin{table*}[t]
\resizebox{0.98\textwidth}{!}{%
\begin{threeparttable}
  \begin{tabular}{lcccc|cccc}
      \toprule
      \multirow{3}{*}{Model} & \multirow{3}{*}{\#Params} &
      \multicolumn{3}{c}{Results on 5s $\uparrow$} &
      \multicolumn{4}{c}{Results on 50s $\uparrow$} \\
    \cmidrule(lr){3-5} \cmidrule(lr){6-9}
     & & Total & Quality & Semantic & Text & Temporal & Dynamic & Framewise \\
     & & Score & Score & Score & Alignment & Quality & Degree & Quality \\
    \midrule
      
    \multicolumn{9}{l}{\textit{Bidirectional models}} \\
    LTX-Video     & 1.9B  & 80.00 & 82.30 & 70.79 & - & - & - & - \\
    Wan2.1        & 1.3B  & 84.67 & 85.69 & 80.60 & - & - & - & - \\
    \midrule
      
    \multicolumn{9}{l}{\textit{Autoregressive models}} \\
    NOVA          & 0.6B  & 80.12 & 80.39 & 79.05 & 24.58 & 86.53 & 31.96 & 34.45 \\
    Pyramid Flow  & 2B    & 81.72 & 84.74 & 69.62 & - & - & - & - \\
    MAGI-1        & 4.5B  & 79.18 & 82.04 & 67.74 & 26.04 & 88.34 & 28.49 & 54.20 \\
    SkyReels-V2   & 1.3B  & 82.67 & 84.70 & 74.53 & 23.73 & 88.78 & 39.15 & 54.13 \\
    CausVid       & 1.3B  & 82.46 & 83.61 & 77.84 & 25.25 & 89.34 & 37.35 & 61.56 \\
    Self-Forcing  & 1.3B  & 83.00 & 83.71 & 80.14 & 24.77 & 88.17 & 34.35 & 61.06 \\
    Self-Forcing++         & 1.3B  & 83.11 & 83.79 & 80.37 & 26.37 & 91.03 & 55.36 & 60.82 \\
    LongLive         & 1.3B  & 83.10 & 83.57 & 81.23 & \textbf{28.08} & 89.29 & 42.40 & \textbf{65.75} \\
    % DySink (Ours)        & 1.3B  & \textbf{83.92} & 84.68 & 80.87 & \textbf{28.39} & \textbf{91.63} & \textbf{63.52} & 64.87 \\
    \textbf{DySink (Ours)}        & 1.3B  & \textbf{84.22} & \textbf{84.82} & \textbf{81.82} & 28.03 & \textbf{92.13} & \textbf{67.70} & 65.34 \\
    \bottomrule
  \end{tabular}
\end{threeparttable}
}
\caption{\textbf{Performance comparison on 5s and 50s videos.}
The 5s setting is evaluated with VBench, while the 50s setting
uses VBench Long. All metrics are higher-is-better, and bold values indicate
the best results among autoregressive models.}
\label{tab:main1} 
\end{table*}

\paragraph{Sink anomaly gate.}

Dynamic frame sinks provide visually relevant historical context, but long-horizon generation can still be affected by sink-collapse attention patterns. Prior analysis shows that, under RoPE, sink-collapse is associated with phase re-alignment to sink frames and inter-head attention homogenization: the failure is not attributed to a single attention head, but occurs when many heads simultaneously assign high attention weights to sink frames, resulting in a degeneracy of attention diversity~\citep{cui2026lol}. Existing methods mitigate this issue via multi-head RoPE jitter, which shifts the base frequencies of different attention heads to break such inter-head homogenization. In contrast, we introduce a lightweight per-layer anomaly gate that uses excessive inter-head consensus as a practical indicator of collapse-prone retrieved context. When a retrieved block exhibits higher head-wise affinity than the local context for an abnormally large fraction of attention heads, the gate removes that block from the layer-wise attention context. This design preserves non-anomalous retrieved blocks, and falls back to the local window only when no retrieved block is retained.

Formally, let ${\mathbf{Q}}^{(j,\ell)} \in \mathbb{R}^{T_q \times H \times d_h}$ denote the queries at layer $\ell$ for the current block $j$, where $T_q$ is the number of query tokens in this block. Let ${\mathbf{K}}_{\mathrm{loc}}^{(j,\ell)}$ denote the corresponding local-window keys. Suppose we retrieve the top-$k$ most relevant memory blocks from the bank, denoted as $\{{\mathbf{K}}_{\mathrm{ret},e}^{(j,\ell)}\}_{e=1}^{k}$. Here, the queries and keys include their RoPE positional encoding, and all
affinities below are computed before the attention softmax. To quantify excessive inter-head consensus, we first compute a head-wise representative query $\bar{\mathbf{q}}_{h}^{(j,\ell)} = \frac{1}{T_q}\sum_{t=1}^{T_q}{\mathbf{Q}}_{t,h}^{(j,\ell)}$. For a key set $\mathbf{K}$, let $|\mathbf{K}|$ denote the number of key tokens,
and let $\mathbf{k}_{r,h}$ be the $r$-th key vector for attention head $h$. We define the average affinity between head $h$ and $\mathbf{K}$ as $a_{h}^{(j,\ell)}(\mathbf{K}) = \frac{1}{|\mathbf{K}|} \sum_{r=1}^{|\mathbf{K}|} \bar{\mathbf{q}}_{h}^{(j,\ell)\top}\,\mathbf{k}_{r,h}$. For each retrieved block $e$, we then measure the fraction of attention heads whose affinity to the retrieved context exceeds that to the local window:
\begin{equation}
\rho_{e}^{(j,\ell)}
= \frac{1}{H}\sum_{h=1}^{H}
  \mathds{1}\!\left[
    a_{h}^{(j,\ell)}\!\bigl({\mathbf{K}}_{\mathrm{ret},e}^{(j,\ell)}\bigr)
    >
    a_{h}^{(j,\ell)}\!\bigl({\mathbf{K}}_{\mathrm{loc}}^{(j,\ell)}\bigr)
  \right].
\end{equation}

We use the local-window affinity as a layer-specific reference rather than
applying an absolute affinity threshold, since attention magnitudes may vary
across layers and generation steps. The consensus ratio therefore measures
how broadly a retrieved block dominates recent context, while remaining
insensitive to isolated high-affinity heads.

For each retrieved block $e$, we apply a block-wise binary gate
$g_{e}^{(j,\ell)}=\mathds{1}\!\left[\rho_{e}^{(j,\ell)} \le
\tau_{\mathrm{gate}}\right]$, where $\tau_{\mathrm{gate}}\in(0,1)$ is a
consensus threshold. Unlike an all-or-nothing filtering rule, the gate is applied
independently to each retrieved block, allowing the layer to suppress only
the anomalous retrieved blocks while retaining the remaining useful historical
context. Let $\mathcal{E}^{(j,\ell)}=\{e \mid g_e^{(j,\ell)}=1\}$ denote the
retained retrieved blocks. The resulting attention keys are

\begin{equation}
\mathbf{K}_{\mathrm{attn}}^{(j,\ell)}
=
\mathrm{Concat}_{\mathrm{token}}
\left(
\left\{
{\mathbf{K}}_{\mathrm{ret},e}^{(j,\ell)}
\right\}_{e \in \mathcal{E}^{(j,\ell)}},
{\mathbf{K}}_{\mathrm{loc}}^{(j,\ell)}
\right).
\end{equation}

When $\mathcal{E}^{(j,\ell)}$ is empty, the attention context falls back to the
local-window keys. The values $\mathbf{V}_{\mathrm{attn}}^{(j,\ell)}$ are
constructed analogously. This gating mechanism adds only a small number of dot-product operations per layer and does not alter the RoPE frequency spectrum or directly modify attention logits.

% \subsection{Training DySink}

% DySink follows the two-stage streaming tuning paradigm of LongLive~\citep{yang2025longlive}, which progressively transitions the model from short-horizon to long-horizon generation.

% \paragraph{Stage~1: Short-horizon video distillation.}

% We initialize the student denoiser from the Wan2.1 model family~\citep{wan2025wan} and perform Distribution Matching Distillation (DMD)~\citep{yin2024onestep} using a high-capacity Wan2.1 teacher denoiser, obtaining a few-step short-video generator~\citep{huang2025self,yang2025longlive}.

% \paragraph{Stage~2: Long-horizon video fine-tuning.}
% Starting from the distilled checkpoint, we fine-tune the generator on long streaming rollouts via LoRA~\citep{hu2022lora, chen2024longlora, yang2025longlive}. To ensure training-inference consistency, the complete DySink pipeline is enabled throughout training. In addition, we adopt two-segment prompt switching~\citep{yang2025longlive}, where the conditioning prompt changes from $p_1$ to $p_2$ at a switching frame uniformly sampled from the valid rollout indices. 

\begin{table*}[t]
  \centering
  \resizebox{0.98\textwidth}{!}{%
  \begin{threeparttable}
    % 将 g 改为标准 c；总列数调整为 9
    \begin{tabular}{lcccc|cccc}
      \toprule
      \multirow{2}{*}{Model} &
      \multicolumn{4}{c}{Results on 75s $\uparrow$} &
      \multicolumn{4}{c}{Results on 100s $\uparrow$} \\
      \cmidrule(lr){2-5} \cmidrule(lr){6-9}
       & Text & Temporal & Dynamic & Framewise & Text & Temporal & Dynamic & Framewise \\
       & Alignment & Quality & Degree & Quality & Alignment & Quality & Degree & Quality \\
      \midrule
      \multicolumn{9}{l}{\textit{Autoregressive models}} \\
      NOVA          & 23.37 & 86.32 & 31.24 & 31.53 & 22.89 & 86.24 & 31.09 & 31.03 \\
      MAGI-1        & 24.95 & 87.89 & 24.82 & 52.04 & 23.75 & 87.62 & 22.21 & 50.90 \\
      SkyReels-V2   & 22.70 & 88.99 & 39.89 & 51.55 & 22.05 & 88.80 & 38.75 & 50.48 \\
      CausVid       & 24.76 & 89.14 & 35.82 & 60.96 & 24.41 & 89.06 & 34.60 & 61.01 \\
      Self-Forcing  & 23.39 & 87.79 & 29.15 & 60.02 & 22.00 & 87.39 & 26.41 & 58.25 \\
      Self-Forcing++         & 26.31 & 91.00 & 55.62 & 60.67 & 26.04 & 90.87 & 54.12 & 60.66 \\
      LongLive  & \textbf{28.21} & 89.24 & 41.96 & \textbf{65.70} & \textbf{28.20} & 89.08 & 40.89 & \textbf{65.72} \\
      \textbf{DySink (Ours)}  & 28.11 & \textbf{92.15} & \textbf{68.05} & 65.24 & 28.08 & \textbf{92.13} & \textbf{68.32} & 65.23 \\
      \bottomrule
    \end{tabular}
  \end{threeparttable}
  }
  \caption{\textbf{Long-horizon performance on 75s and 100s videos.}
Results are evaluated with VBench Long on the same 128-prompt benchmark.
All metrics are higher-is-better, and bold values indicate the best results among autoregressive models.}
  \label{tab:main2} % \label 已移至 \caption 正下方
\end{table*}

\section{Experiments}
\label{sec:experiment}

\subsection{Implementation Details}
\label{sec:implementation}

We build DySink on Wan2.1-T2V-1.3B~\citep{wan2025wan}, which can generate
5-second videos at 16 FPS and a resolution of $832\times480$. Following the
two-stage streaming tuning paradigm of LongLive~\citep{yang2025longlive},
we first perform Distribution Matching Distillation
(DMD)~\citep{yin2024onestep} using Wan2.1-T2V-14B as the teacher, obtaining
a few-step causal short-video generator. We then fine-tune the distilled
checkpoint on streaming rollouts of up to 60 seconds via
LoRA~\citep{hu2022lora,chen2024longlora}. The complete DySink pipeline is
enabled during Stage~2 training to maintain training--inference consistency.

% Following LongLive, we additionally use two-segment prompt switching as a
% training augmentation, with the switching point uniformly sampled from the
% valid rollout indices.

We use PE-Core-S~\citep{bolya2025pe} as the frozen visual encoder and
set $\tau_{\mathrm{dedup}}=0.95$, $\tau_{\mathrm{gate}}=0.8$, and $k=2$.
Each memory block contains three latent frames, and the local window contains
three blocks. With $k=2$, the default configuration therefore uses six retrieved and nine
local latent frames, corresponding to D6L9 under a 15-frame active-context
budget. Both training stages use 3,000 optimization steps. Additional
optimization details are provided in the supplementary material.

\subsection{Evaluation}

\paragraph{Evaluation Metrics} 
We evaluate our model under two settings to assess both generation quality and temporal robustness. 
(1) \textbf{Short-horizon generation (5s)}: Following the VBench~\citep{huang2024vbench}, we use 946 prompts across 16 dimensions, reporting the \textit{Total Score}, \textit{Quality Score}, and \textit{Semantic Score} to quantify visual fidelity and semantic alignment. 
(2) \textbf{Long-horizon generation (50s, 75s, 100s)}: To examine the capacity for extended generation, we adopt a prompt set of 128 samples from MovieGen~\citep{polyak2024movie}, following the experimental setup of CausVid~\citep{yin2025causvid} and Self-Forcing++~\citep{cui2025self}. Performance in
this setting is assessed with VBench Long, with evaluation focused on \textit{Text Alignment}, \textit{Temporal Quality}, \textit{Dynamic Degree}, and \textit{Framewise Quality}. 

To explicitly evaluate sink-collapse-like history returns, we additionally report the NoRepetition Score in the ablation study. A history-return event is detected when a frame both changes abruptly from its predecessor and closely matches a frame generated at least 15 seconds earlier. The score is the percentage of videos containing no detected event, with higher values indicating stronger resistance to long-horizon recurrence. The implementation details are provided in the supplementary material. Together, these two settings provide a complementary evaluation for short-horizon and long-horizon generation quality.

\paragraph{Baseline methods} We compare against several existing approaches, including NOVA~\citep{deng2024nova}, Pyramid Flow~\citep{jin2024pyramidal}, SkyReels-V2-1.3B~\citep{chen2025skyreels}, MAGI-1-4.5B~\citep{teng2025magi}. We also include the similar autoregressive methods CausVid~\citep{yin2025causvid}, Self-Forcing~\citep{huang2025self}, Self-Forcing++~\citep{cui2025self} and LongLive~\citep{yang2025longlive}, all of which are 1.3B distilled few-step generators. For reference, two bidirectional models, LTX-Video~\citep{hacohen2025ltx} and Wan2.1-1.3B~\citep{wan2025wan}, are also evaluated. Unless otherwise noted, the baseline results in Tables~\ref{tab:main1} and~\ref{tab:main2} are taken from the benchmark results provided by Self-Forcing++; for LongLive, we reproduce the results using the released weights and the corresponding hyperparameters. Self-Forcing++ represents a strong local-window autoregressive baseline, while LongLive provides the closest static-sink counterpart based on the same 1.3B model.

\paragraph{Short-Horizon Generation (5s).}
As shown in Table~\ref{tab:main1}, DySink achieves the highest Total Score, Quality Score, and Semantic Score among autoregressive methods, outperforming strong baselines such as Self-Forcing++ and LongLive. Since 5-second generation does not yet require substantial long-range retrieval, these results primarily verify that the Stage~2 long-horizon fine-tuning and memory-control pipeline do not compromise
the short-horizon fidelity or semantic alignment of the base generator.

\begin{table*}[t]
    \centering
    \begingroup
    \renewcommand{\arraystretch}{1.0}
    \setlength{\tabcolsep}{4pt}
    \normalsize

    \begin{tabular}{lccccccc}
        \toprule
        \textbf{Method} &
        \makecell{\textbf{Text}\\\textbf{Alignment} $\uparrow$} &
        \makecell{\textbf{Background}\\\textbf{Consistency} $\uparrow$} &
        \makecell{\textbf{Subject}\\\textbf{Consistency} $\uparrow$} &
        \makecell{\textbf{Motion}\\\textbf{Smoothness} $\uparrow$} &
        \makecell{\textbf{Dynamic}\\\textbf{Degree} $\uparrow$} &
        \makecell{\textbf{Framewise}\\\textbf{Quality} $\uparrow$} &
        \makecell{\textbf{NoRepetition}\\\textbf{Score} $\uparrow$} \\
        \midrule

        S0L15
        & 27.76 & 94.73 & 96.83 & 97.95 & 57.27 & 64.88 & 100.00 \\
        \addlinespace[2pt]

        S3L12 w/o G
        & 28.44 & 95.37 & 97.15 & 98.21 & 57.39 & 66.41 & 26.56 \\
        \addlinespace[2pt]

        \makecell[l]{S3L12 w/ G\\$(\tau_{\mathrm{gate}}=0.8)$}
        & 28.30 & 95.27 & 97.16 & 98.21 & 56.52 & 66.22 & 89.06 \\
        \addlinespace[2pt]

        D3L12 w/o G
        & 27.81 & 95.13 & 96.90 & 97.91 & 54.00 & 66.05 & 75.78 \\
        \addlinespace[2pt]

        \makecell[l]{D3L12 w/ G\\$(\tau_{\mathrm{gate}}=0.8)$}
        & 27.72 & 95.06 & 96.80 & 97.84 & 56.49 & 65.74 & 95.31 \\
        \addlinespace[2pt]

        S6L9 w/o G
        & 28.53 & 95.39 & 97.05 & 98.51 & 52.28 & 65.91 & 18.75 \\
        \addlinespace[2pt]

        \makecell[l]{S6L9 w/ G\\$(\tau_{\mathrm{gate}}=0.8)$}
        & 28.36 & 95.01 & 96.96 & 98.60 & 50.18 & 64.32 & 55.47 \\
        \addlinespace[2pt]

        \makecell[l]{S6L9 w/ G\\$(\tau_{\mathrm{gate}}=0.6)$}
        & 28.19 & 94.91 & 96.80 & 98.52 & 52.73 & 63.93 & 85.16 \\
        \addlinespace[2pt]

        D6L9 w/o G
        & 28.20 & 94.87 & 96.23 & 97.56 & 67.40 & 65.78 & 31.25 \\
        \addlinespace[2pt]

        \makecell[l]{D6L9 w/ G\\$(\tau_{\mathrm{gate}}=0.8)$}
        & 28.03 & 94.86 & 96.19 & 97.56 & 67.70 & 65.34 & 85.16 \\

        \bottomrule
    \end{tabular}

    \endgroup
    \caption{Ablation study of historical-context composition and sink anomaly
gating on 50-second video generation. }
\label{tab:ablation}
\end{table*}

% Here, $S$, $D$, $L$, and $G$ denote static sinks, dynamic sinks, local frames, and the anomaly gate, respectively. 
% All models use the same Stage~1
% initialization, Stage~2 training budget, and 15-frame historical-context
% budget. The w/o G results are obtained by disabling the gate for the
% corresponding model weights.

\paragraph{Long-Horizon Generation (50s / 75s / 100s).}
The advantages of DySink become more pronounced in long-horizon generation.
Across the 50s, 75s, and 100s settings in Tables~\ref{tab:main1}
and~\ref{tab:main2}, DySink consistently achieves the highest measured
\emph{Temporal Quality} and \emph{Dynamic Degree} among autoregressive
baselines. Compared with Self-Forcing++, DySink improves \emph{Dynamic Degree}
by +12.34 / +12.43 / +14.20 points on 50s / 75s / 100s videos, respectively,
while also improving \emph{Temporal Quality} by +1.10 / +1.15 / +1.26 points,
\emph{Text Alignment} by +1.66 / +1.80 / +2.04 points, and
\emph{Framewise Quality} by +4.52 / +4.57 / +4.57 points. These gains
across temporal, semantic, and frame-level metrics indicate that DySink supports
richer long-horizon evolution while preserving coherence, text alignment, and
visual quality.

Compared with LongLive, which relies on static sink frames for long-range
stabilization, DySink achieves substantially higher \emph{Dynamic Degree}
(+25.30 / +26.09 / +27.43 on 50s / 75s / 100s videos) and higher
\emph{Temporal Quality} (+2.84 / +2.91 / +3.05), while maintaining comparable
\emph{Text Alignment} and \emph{Framewise Quality}.

\subsection{Ablation Study}
\label{sec:ablation}

\paragraph{Experimental setup.}
We study the effects of historical-context allocation, sink selection, and the
sink anomaly gate on 50-second video generation. All trained sink-allocation variants are initialized from the same checkpoint and further optimized under identical settings. For the
w/o G variants, we use the corresponding trained weights and disable
the anomaly gate only during inference. The total historical-context budget is fixed to 15 latent frames. \emph{S0L15} uses only a 15-frame local window;
\emph{S3L12} and \emph{S6L9} allocate three and six frames to static sinks,
respectively; and \emph{D3L12} and \emph{D6L9} replace the static sinks with
dynamically retrieved frames. Unless otherwise specified, the sink-based
models use the anomaly gate with $\tau_{\mathrm{gate}}=0.8$ during training
and evaluation. The corresponding \emph{w/o G} results are obtained by
disabling the gate at inference for the same model weights, providing a paired
evaluation of its inference-time effect. Here, $S$, $D$, $L$, and $G$ denote
static sinks, dynamic sinks, local frames, and the anomaly gate, respectively.

\paragraph{Results.}
Table~\ref{tab:ablation} reveals three main observations. First, the anomaly
gate primarily improves resistance to history-return failures rather than
serving as a generic quality booster. With $\tau_{\mathrm{gate}}=0.8$, enabling
the gate increases the \emph{NoRepetition Score} by 19.53--62.50 points across
sink-based variants, while changing text alignment, consistency, smoothness,
and framewise quality only marginally. For instance, \emph{D6L9} improves from
31.25 to 85.16 in \emph{NoRepetition Score}, with nearly unchanged
\emph{Dynamic Degree} (67.40 vs.\ 67.70). The slight decreases in
consistency-oriented metrics are expected, since these metrics can favor
conservative generations with similar content across frames and
may therefore reward videos that evolve less or repeatedly revisit earlier
states. By suppressing over-dominant historical blocks, the gate reduces such
over-anchoring and allows stronger visual evolution, but may occasionally remove
stabilizing context useful for preserving fine-grained background or subject
appearance. 

% Overall, the large gain in \emph{NoRepetition Score} with only
% minor changes in conventional metrics supports the role of the gate as a
% targeted mechanism for suppressing collapse-prone retrieved context.

Second, static sinks become more vulnerable as more budget is assigned to them.
Without gating, increasing static sinks from \emph{S3L12} to \emph{S6L9}
reduces the \emph{NoRepetition Score} from 26.56 to 18.75, suggesting that
persistent early-frame anchors can amplify history-return behavior. The gate
partially alleviates this issue, but even with a stricter threshold
($\tau_{\mathrm{gate}}=0.6$), \emph{S6L9} remains substantially less dynamic
than the dynamic-sink counterpart.

Third, dynamic retrieval improves long-horizon evolution under the same
context budget. Comparing \emph{D6L9} and \emph{S6L9}, dynamic sinks achieve a
much higher \emph{Dynamic Degree} (67.70 vs.\ 50.18) and a higher
\emph{NoRepetition Score} (85.16 vs.\ 55.47) with comparable text alignment and
framewise quality. Although the local-only \emph{S0L15} obtains the highest
\emph{NoRepetition Score}, it lacks non-local memory and yields lower dynamics
and framewise quality than \emph{D6L9}. We therefore choose \emph{D6L9} with
$\tau_{\mathrm{gate}}=0.8$ as the final configuration, as it offers the best
balance between active visual evolution, visual quality, and repetition
resistance.

\section{Conclusion}
In this work, we revisited the design of bounded historical context for autoregressive long video generation. Motivated by the limited adaptability of persistent early-frame sinks, we introduced DySink, which retrieves relevant historical context from a compact memory bank and uses excessive inter-head consensus as a gating signal to suppress anomalous retrieved context. Across 50--100-second videos, DySink achieves the highest measured temporal quality among the evaluated autoregressive baselines, while retaining competitive text alignment and framewise quality. These results show that DySink offers a favorable balance between long-range
temporal coherence and active visual evolution under a fixed active-context budget.

\bibliography{aaai2027}

\end{document}